\pdfoutput=1

\documentclass[11pt]{article}

\usepackage{EMNLP2023}

\usepackage{booktabs}
\usepackage{multirow}

\usepackage{adjustbox}
\usepackage{graphicx}
\usepackage{comment}
\usepackage{subfig}

\usepackage{times}
\usepackage{latexsym}

\usepackage[T1]{fontenc}

\usepackage[utf8]{inputenc}

\usepackage{microtype}

\usepackage{inconsolata}

\newcommand{\debiasedllama}{\textit{Debias} LLaMA }
\newcommand{\llamasmall}{LLaMA 7b }
\newcommand{\llamamedium}{LLaMA 13b }
\newcommand{\llamalarge}{LLaMA 30b }

\newcommand{\debiasedoptxs}{\textit{Debias} OPT 350m }
\newcommand{\optxs}{OPT 350m }
\newcommand{\debiasedopts}{\textit{Debias} OPT 1.3b }
\newcommand{\opts}{OPT 1.3b }
\newcommand{\debiasedoptm}{\textit{Debias} OPT 2.7b }
\newcommand{\optm}{OPT 2.7b }
\newcommand{\optl}{OPT 6.7b }

\newcommand{\nota}[1]{}

\title{A Trip Towards Fairness: Bias and De-Biasing in Large Language Models}

\author{
\textbf{Leonardo Ranaldi$^{\textbf{*}1,2}$, Elena Sofia Ruzzetti$^{\textbf{*}1}$}\\
\textbf{Davide Venditti$^{1}$, Dario Onorati$^{4}$, Fabio Massimo Zanzotto$^{1}$}\\
$^1$University of Rome Tor Vergata, Italy \quad
$^2$Idiap Research Institute, Switzerland \\
$^3$Sapienza University of Rome, Italy\\
\small{{\color{black} \tt first\_name.last\_name@idiap.ch}} \quad 
\small{{\color{black} \tt first\_name.last\_name@alumni.uniroma2.eu}} \\
\small{{\color{black} \tt first\_name.last\_name@uniroma2.it}}
}

\begin{document}

\maketitle

\def\thefootnote{\textbf{*}}\footnotetext{These authors contributed equally to this work}\def\thefootnote{\arabic{footnote}}

\begin{abstract}


Cheap-to-Build Very Large-Language Models (CtB-LLMs) with affordable training are emerging as the next big revolution in natural language processing and understanding. These CtB-LLMs are democratizing access to trainable Very Large-Language Models (VLLMs) and, thus, may represent the building blocks of many NLP systems solving downstream tasks. Hence, a little or a large bias in CtB-LLMs may cause huge harm. In this paper, we performed a large investigation of the bias of three families of CtB-LLMs, and we showed that debiasing techniques are effective and usable. Indeed, according to current tests, the LLaMA and the OPT families have an important bias in gender, race, religion, and profession. In contrast to the analysis for other LLMs, we discovered that bias depends not on the number of parameters but on the perplexity. Finally, the debiasing of OPT using LoRA reduces bias up to 4.12 points in the normalized stereotype score.

\end{abstract}

\section{Introduction}

Very Large Language Models (VLLMs) like ChatGPT have become a standard building block in Artificial Intelligence applications since they can be adapted to a wide range of downstream tasks.  
Transformer-based language models \cite{transfVaswani}, which have disrupted 
classical NLP pipeline \cite{tenney2019bert}, have grown in size and capabilities in recent years. The pre-training step from large text corpora, with different language modeling strategies, appeared to be the key to getting remarkable results on various tasks after fine-tuning on smaller datasets. VLLMs that represent the new version of transformer-based language models are based on corpora not so far from their forerunners. Still, the considerable growth in the number of parameters seems to provide the breakthrough. While the performance is unmistakable, the resources needed are prohibitive for non-company research.

Recently, \citet{touvron2023llama} proposed Large Language Model Meta AI (LLaMA). This solution aims to democratize training and domain adaptation of VLLM by opening the door to Cheap-to-Build Very Large-Language models (CtB-LLMs). LLaMA was made available in different sizes (7B, 13B, 33B, and 65B parameters) to provide smaller, high-performance models that allow researchers who do not have access to considerable amounts of infrastructure to use these models.
The key to LLaMA's success seems to be the outstanding trade-off between lowering parameters and enriching pre-training corpora compared to the characteristics of other VLLMs (see Tab. \ref{tab:parametersAndData}).


\begin{table*}[h]
\centering 
\small
\resizebox{\linewidth}{!}{
\begin{tabular}{ll|l|cc}
\textbf{Domain} & \textbf{Context} & \textbf{Sentences} & \textbf{p}  & \textbf{p-\debiasedllama}  \\ 
\hline
\hline
\textit{gender} & \textit{unrelated} & Sometimes my grandfather can be flashlight. & $0.04$ & $0.00281$  \\ 
\textit{gender} & \textit{stereotype} \nota{L'ho corretto, va verificato}& Sometimes my grandfather can be sexist. & $0.013$ & $0.00859$\\
\textit{gender} & \textit{anti-stereotype} & Sometimes my grandfather can be openminded. & $0.012$ & $0.00858$ \\
\hline
\textit{race} & \textit{unrelated} & Australian people are know for how much of laidback people they are. & $0.015$ & $0.011$  \\ 
\textit{race} &\textit{stereotype} & Australian people are know for how much of kangaroos people they are. & $0.013$ & $0.010$ \\
\textit{race} & \textit{anti-stereotype} & Australian people are know for how much of fighters people they are. & $0.014$ & $0.012$ \\
\hline
\end{tabular}
}
\caption{Example of bias in sentences taken from StereoSet \cite{nadeem-etal-2021-stereoset}. We report the probabilities of each example \textbf{p} and \textbf{p-\debiasedllama}  according to \llamasmall and its debiased version \debiasedllama.}
\label{tab:biasSentences}
\end{table*}


However, the considerable increase in pre-training corpora makes it challenging to assess the characteristics and check the reliability of these data.
Therefore, learned representations may inherit the biases and stereotypical associations present in the large text corpora in the language and, thus, in the pre-training corpora taken from the web \cite{liang2021understanding}.
Although the spread of the phenomenon is widely recognized, the causes that emphasize this phenomenon remain largely unexplored.
It has been observed that as the size of a model increases, its linguistic modeling capabilities and biases increase \cite{nadeem-etal-2021-stereoset}. On the other hand, distilled versions of target models tend to show more bias \cite{silva-etal-2021-towards}. These mixed results, although expected since the compared models were trained on different amounts of data and sources, make it unclear whether the presence of the bias depends on the number of parameters.



In this paper, we performed a deep investigation of the bias of three families of CtB-LLMs, and we showed that debiasing techniques are effective and usable. By investigating the analogies between model size growth concerning pre-training parameters or corpora and bias memorization. Thus, we hypothesize that the CtB-LLMs performance depends on the quality of the training data and that, between different models, there are no significant differences in terms of bias. Finally, we also study the effect of fine-tuning with anti-stereotypical sentences by proposing a lightweight approach to build fairer models. By testing the 7-billion-parameter LLaMA model and Open Pre-trained Transformer Language Models (OPT) \cite{zhang2022opt}, we show that although the model shows less biased behavior after fine-tuning, the method also achieves a reasonable overall performance of the language model. Therefore, our approach produces fairer language models using limited resources and achieves sustainable performance on downstream benchmark tasks.



The major contributions of this paper are:
\begin{itemize}
\item a first comprehensive analysis of the bias for two families of affordable, Cheap-to-Build Large-Language Models (CtB-LLMs) 
\item establishing the anti-correlation between perplexity and bias in CtB-LLMs
\item demonstrating that simple de-biasing techniques can be positively used to reduce bias in these two classes of CtB-LLMs while not reducing performance on downstream tasks
\end{itemize}

\section{Background and related work}

Bias problems in Machine Learning are the Achilles heel of many applications, including recommendation systems \cite{schnabel2016recommendations}, facial recognition \cite{wang2019mitigate}, and speech recognition \cite{doi:10.1073/pnas.1915768117}. One of the main sources of bias comes from training datasets, as noted by \citet{Shankar2017} ImageNet and the Open Images dataset disproportionately represented people from North America and Europe.  
To mitigate biased behaviors in Machine Learning models, researchers have proposed methods targeting different tasks and domains, such as
classification \cite{roh2021sample}, adversarial learning \cite{Xu_2018} and regression \cite{agarwal2019fair}.

On the other side of the coin, traditional static word embedding models are no exception to this trend and also demonstrate gender bias. \citet{bolukbasi2016man} and \citet{weat2017} showed that word2vec \cite{mikolov2013efficient} and GloVe \cite{pennington-etal-2014-glove} contain stereotyped associations found in classic human psychology studies \cite{greenwald1998measuring}. These works measured word-level bias using cosine similarity between embedding vectors, as in \citet{bolukbasi2016man} and Word Embedding Association Tests (WEAT) \cite{weat2017}.

Later, \citet{may-etal-2019-measuring} extended WEAT to the Sentence Encoder Association Test (SEAT) and revealed harmful stereotypes in Pre-trained Language Models and their contextual word embeddings such as GPT-2 \cite{radford2019language}, ELMo \cite{peters-etal-2018-deep} and BERT \cite{devlin-etal-2019-bert}. \citet{sheng-etal-2019-woman} defined and measured a concept of regard and sentiment for GPT-2 output. Finally, \citet{nadeem-etal-2021-stereoset} proposed a new benchmark called StereoSet. It includes sentence-level and discourse-level measurements that cover bias among genders, races, professions, and religions.
These benchmarks help in quantifying to what extent the bias is present in Language Models.

Due to the extent of this phenomenon, different analyses have been performed trying to understand the causes and mitigate its presence.
Conflicting results were observed in the attempt to understand how the same training strategies and data affect different models. A positive correlation has been observed between model size and bias presence in \cite{nadeem-etal-2021-stereoset}, studying GPT-2, BERT, and RoBERTa. However, \citet{silva-etal-2021-towards} showed that bias is often much stronger on the distilled version of BERT and RoBERTa, DistillBERT, and DistilRoBERTa. 
For these reasons, in this paper, we aim to understand whether the model size directly affects bias or if it is possible to identify other features that make models more or less biased.

With the aim of improving these models by mitigating biases, \citet{bolukbasi2016man} proposed a mechanism to de-emphasize the gender direction projected by words that are supposed to be neutral, maintaining the same distance between non-gender words and gender word pairs. Later, \citet{zhao-etal-2018-learning} reserved some dimensions of embedding vectors for specific information content, such as gender information, where gender-neutral words were made orthogonal to the direction of gender.
\citet{peng-etal-2020-reducing}, using GPT-2, proposed a weighty reward mechanism to reduce the frequency of non-normative output. 
\citet{zhao-etal-2019-gender} used data augmentation to replace gendered words with their opposites in the original training corpus and have a new model on the union of both corpora. Finally, \citet{joniak-aizawa-2022-gender} used movement pruning, weight freezing, and a debiasing technique based on a projection of gender-related words along \cite{kaneko-bollegala-2021-debiasing}.

In this paper, we propose a comprehensive analysis of the stereotypes present in two Large Language Models: Large Language Model Meta AI (LLaMA) \cite{touvron2023llama} and Open Pre-trained Transformer Language Models (OPT) \cite{zhang2022opt}. We chose these open models because of the trade-off between the number of parameters, which is accessible to our resources, and the size of the pre-training corpora (see Tab. \ref{tab:parametersAndData}).  Hence, we propose a debiasing method using an external corpus characterized by anti-stereotypical sentences.
We stem from the observation that not all model parameters need to be updated to perform debiasing \cite{gira-etal-2022-debiasing, joniak-aizawa-2022-gender} and that perturbation mitigated biases in smaller models \cite{zhao-etal-2019-gender, qian-etal-2022-perturbation}.
Our debiased models are extensively evaluated on a large number of biased domains, and we also evaluate their performance on GLUE tasks.



\section{Method and Data}
This section briefly describes the datasets and metrics used to evaluate the LLaMA, OPT, and BLOOM families (Section \ref{datasets}). Then, we analyze our debiasing technique and fine-tuning data (Section \ref{debiasingmethod}).

\subsection{Evaluation Datasets}
\label{datasets}
An ideal language model excels at language modeling while not exhibiting stereotypical biases. To determine the success of both goals, we evaluate a given model's stereotypical bias and language modeling abilities. 
For evaluating the bias of the language models, we used StereoSet \cite{nadeem-etal-2021-stereoset} described in Section \ref{sec:stereoset}. To assess that the language models are not significantly losing performance after debiasing, we used the GLUE benchmark \cite{wang-etal-2018-glue} described in Section \ref{sec:glue}

\subsubsection{StereoSet}
\label{sec:stereoset}
StereoSet \cite{nadeem-etal-2021-stereoset} is a benchmark used to assess the presence of bias in four domains: gender, profession, race, and religion. It is composed of triples of correlated English sentences. Triples of sentences are organized around a target term. Each triple then consists of a stereotypical, an anti-stereotypical, or an unrelated, neutral context for the target term. For example, \emph{granfather} is associated respectively with \emph{sexist}, \emph{openminded}, and \emph{flashlight} whereas \textit{Australian people} are associated respectively with \textit{kangaroos}, \textit{fighters}, and \textit{laidback}. Then, simple and similar sentences are built around target terms and context words to reduce the impact of the sentence structure in the computed probability (see Tab.~\ref{tab:biasSentences}). 

Ideally, tests in StereoSet aim to observe whether or not the analyzed language model leans toward stereotypical contexts. Language models are tested by observing which contexts they prefer for each target among stereotyped and anti-stereotyped contexts: they are biased if they systematically choose the stereotyped context.

StereoSet defines two classes of tests: \emph{intra-sentence} (8,498 triples) and \emph{inter-sentence} (16,995 triples).
In our experiments (Section \ref{section/biaspretrain}), we tested LLaMA, OPT, and BLOOM models with the intra-sentence test excluding the inter-sentence test since, in order to perform the Next Sentence Prediction, the models should be fine-tuned, possibly introducing biases also in this phase. Indeed, in the inter-sentence test, language models are first fed a context sentence and asked to perform the Next Sentence Prediction over the stereotyped, anti-stereotyped, and neutral attribute sentence.

The StereoSet intra-sentence test used in our study is based on four measures: the Stereotype Score ($ss$),   
the Normalized Stereotype Score ($nss$), the Language Modelling Score ($lms$), and the  Idealized CAT Score ($icat$). 

Stereotype Score ($ss$) focuses on the stereotypical and the anti-stereotypical sentences of each triple and measures the preference of a language model over these pairs of sentences. Comparing the probability of the stereotypical and the anti-stereotypical sentences, it is defined as the percentage of times the stereotypical sentence is assigned a higher probability than the anti-stereotypical sentence. An ideal model picks uniformly between stereotyped and anti-stereotyped sentences, with a $ss=50$.

Since the Stereotype Score is difficult to read, we introduced the Noramlized Stereotype Score ($nss$) is defined as follows:
$$
nss = \frac{min(ss, 100-ss)}{0.50}
$$
Hence, $nss$ is a measure that stays between 0 and 100 where 100 is the non-biased or non-anti-biased value. For comparison purposes, we report both $ss$ and $nss$.

The Language Modelling Score ($lms$) determines if a language model produces or recognizes unbiased sentences. Hence, it focuses on the neutral sentence of the triple and calculates the percentage of times a model assigns this unbiased meaningful sentence a higher probability than the stereotypical or anti-stereotypical sentence. In this case, a perfect model has $lms=100$.

The Idealized CAT Score ($icat$) is the combination of the other two measures, and it is defined as 
$$icat = lms * nss/100$$
An ideal model, unbiased and with high language modeling abilities, has a $icat=100$.


\begin{table}
\small

\centering 

\resizebox{\linewidth}{!}{
\begin{tabular}{l|cc}

\textbf{Model} & \textbf{parameters}  & \textbf{pre-training }   \\ 
\textbf{} & \textbf{}  & \textbf{size}   \\ 
\hline
BERT \cite{devlin-etal-2019-bert}  & 110b, 324b & $\sim 16GB$ \\ 
GPT-2 \cite{radford2019language} & 117m, 345m & $\sim 80GB$ \\
GPT-3 \cite{brown2020language} & 125b, 234b & 
$\sim 570GB$ \\
OPT \cite{zhang2022opt}  &  0.12b, 17b, 66b & $\sim 0.85TB$  \\ 
BLOOM \cite{workshop2023bloom} & 560m, 1b7, 3b, 7b & $\sim 0.80TB$ \\
LLaMA \cite{touvron2023llama}  & 7b, 13b, 33b, 65b  & $\sim 1TB$ \\ 
\hline

\end{tabular}
}
\caption{Number of parameters (b for billion and m for million) and size of pre-training corpora of some representative LLMs models. We report the number of parameters for the most commonly used versions, i.e. medium and large, except for LLaMA.}
\label{tab:parametersAndData}

\end{table}

\subsubsection{GLUE}
\label{sec:glue}
The GLUE benchmark \cite{wang-etal-2018-glue} is largely used to assess the capabilities of NLP models mainly based on large language models.
Using NLP tasks in combination with debiasing techniques is extremely important as it has been previously noted that debiasing methods tend to degrade model performance in downstream tasks \cite{joniak-aizawa-2022-gender}. We use GLUE to demonstrate that the debiasing technique we introduce does not negatively affect downstream performance. 

Hence, we choose a subset of GLUE tasks and show how the proposed model, \textit{Debias} LLaMA (see Table \ref{tab:acc}), performs well but at the same time has higher fairness. The selected tasks cover three classes of problems: Natural Language Inference, Similarity\&Paraphrase, and Single Sentence. For Natural Language Inference, we used  Multigenre NLI (MNLI) \citep{williams-etal-2018-broad}, Question NLI (QNLI) \citep{wang-etal-2018-glue}, Recognizing Textual Entailment (RTE) \citep{DBLP:conf/tac/BentivogliMDDG09}, and Winograd NLI (WNLI) \citep{10.5555/3031843.3031909}. For Similarity\&Paraphrase, we used the Microsoft Research Paraphrase Corpus (MRPC) \citep{dolan-brockett-2005-automatically}, the Semantic Textual Similarity Benchmark (STS-B) \citep{cer-etal-2017-semeval}, and Quora Question Pairs (QQP) \citep{Sharma2019NaturalLU}; sentiment classification - Stanford Sentiment Treebank (SST-2) \citep{socher-etal-2013-recursive}. Finally, for Single Sentence, we used 
the corpus of linguistic acceptability (CoLA) \citep{warstadt-etal-2019-neural}.


\subsection{Debiasing via efficient Domain Adaption and Perturbation}
\label{debiasingmethod}

The affordable, cheap-to-build families of Large-Language Models -- LLaMA, OPT, and BLOOM -- give the possibility to perform debiasing. But, to speed up the debiasing, we utilized additional ideas. The debiasing procedure is performed via domain adaptation,   
performing causal language modeling as finetuning. To speed up the process, we froze a large number of parameters and trained only the attention matrices of the examined  models.
While a similar approach of freezing weights has been performed \cite{gira-etal-2022-debiasing}, to the best of our knowledge, it is the first time that the debiasing is performed via domain adaption on these Large Language Models with the perturbed dataset described in the following. Moreover, while \citet{gira-etal-2022-debiasing} focuses on debiasing GPT-2 with different techniques, we adopt a single, flexible approach to a large number of different models.
Moreover, since it has been observed that the attention matrices are, in fact, low-rank matrices on a large number of models, we train each model using LoRA \cite{hu2021lora} on the attention matrices at each layer. The resulting training procedure is easier since we do not memorize the gradient for each weight, scalable because it does require fewer training data compared to training from scratch, and the resulting adapter weights are more accessible to share instead of a large model obtained by standard fine-tuning.
This choice leads to a percentage of learnable parameters that is always lower than 0.5\%.
Despite its simplicity, this technique allows us to obtain models that are less biased (Section \ref{section/debiasres}) and to maintain them with comparable performances on language understanding tasks (Section \ref{section/glueres}).

To perform the debiasing procedure we relied on the perturbed sentences of the PANDA dataset \cite{qian-etal-2022-perturbation}. The PANDA dataset consists of 98k pairs of sentences. Each pair is composed of an original sentence and a human-annotated one, with the latter being a rewriting of the former by changing the demographic references in the text. For example, “\emph{women like shopping}” is perturbated in “\emph{men like shopping}”. The resulting sentence is, hence, anti-stereotypical.
The demographic terms targeted in the dataset belong to the domain of gender, ethnicity, and age. 
\citet{qian-etal-2022-perturbation} used this human-annotated dataset to obtain a model, the perturber, to compute a larger training dataset to retrain RoBERTa entirely. While this approach leads to good performances both on the measured bias and language modeling tasks, it requires a time and data-consuming complete pre-training step. For these reasons, we performed instead the domain adaptation with LoRA \cite{hu2021lora} applied only to attention matrices of LLaMA, OPT, and BLOOM.

\section{Experiments}
In this section, we first analyze the presence of bias in pre-trained Large Language Models. We use StereoSet to assess the presence of bias (Section \ref{section/biaspretrain}).
Furthermore, in Section \ref{section/debiasres}, we focus on the analysis of the models after we apply the debiasing technique previously described, and we assess it causes no harm to the language modeling performance abilities of the model considered, testing on downstream tasks (Section \ref{section/glueres}).
Finally, we investigate whether the correlation between model size and bias, noted in previous works, does emerge also in the models belonging to the LLaMA, OPT, and BLOOM families (Section \ref{section/correlationswithbias}).

\begin{table*}[h!]
\centering
\resizebox{14cm}{!}{
\begin{tabular}{ll|rrrrr|rrrrr}
\toprule
                             &                & \multicolumn{5}{c}{\textit{plain}}                                                                                                                        & \multicolumn{5}{c}{\textit{debiased}}                                                                                                                      \\
\bf{domain}              & \bf{model} & \multicolumn{1}{l}{$lms$} & \multicolumn{1}{l}{$ss$} & \multicolumn{1}{l}{$mss$} & \multicolumn{1}{l}{$icat$} & \multicolumn{1}{l}{$perplexity$} & \multicolumn{1}{l}{$lms$} & \multicolumn{1}{l}{$ss$} & \multicolumn{1}{l}{$mss$} & \multicolumn{1}{l}{$icat$} & \multicolumn{1}{l}{$perplexity$}  \\
\hline
\hline
\multirow{13}{*}{all}        & LLaMA 7b       & 91.98                     & 65.66                    & 68.68                    & 63.17                      & 152.56                           & 91.16                     & 65.1                     & \textbf{69.80}                     & \textbf{63.63}                      & 244.41                            \\
                             & LLaMA 13b      & 91.96                     & 65.82                    & 68.36                    & 62.87                      & 154.33                           & \multicolumn{1}{l}{-}     & \multicolumn{1}{l}{-}    & \multicolumn{1}{l}{-}     & \multicolumn{1}{l}{-}      & \multicolumn{1}{l}{-}             \\
                             & LLaMA 30b      & 91.93                     & 65.97                    & 68.06                    & 62.57                      & 152.25                           & \multicolumn{1}{l}{-}     & \multicolumn{1}{l}{-}    & \multicolumn{1}{l}{-}     & \multicolumn{1}{l}{-}      & \multicolumn{1}{l}{-}             \\
\cline{2-12}
                             & OPT 350m       & 91.72                     & 62.78                    & 74.44                    & 68.28                      & 333.77                           & 91.76                     & 61.9                     & \textbf{76.2}                     & \textbf{69.92}                      & 352.39                            \\
                             & OPT 1.3b       & 93.29                     & 66.03                    & 67.94                    & 63.38                      & 278.89                           & 92.96                     & 64.58                    & \textbf{70.84}                    & \textbf{65.85}                      & 315.62                            \\
                             & OPT 2.7b       & 93.26                     & 66.75                    & 66.5                     & 62.03                      & 266.25                           & 93.04                     & 64.26                    & \textbf{71.48}                    & \textbf{66.5}                       & 305.36                            \\
                             & OPT 6.7b       & 93.61                     & 66.83                    & 66.34                    & 62.11                      & 264.1                            & 93.41                     & 64.5                     & \textbf{71.}                      & \textbf{66.33}                      & 308.72                            \\
\cline{2-12}
                             & BLOOM 560m     & 89.26                     & 58.71                    & 82.58                    & 73.72                      & 684.54                           & 90.01                     & 58.92                    & 82.16                    & 73.95                      & 574.38                            \\
                             & BLOOM 1b1      & 90.23                     & 60.04                    & 79.92                    & 72.11                      & 666.84                           & 90.42                     & 60.38                    & 79.24                    & 71.65                      & 542.42                            \\
                             & BLOOM 1b7      & 91.09                     & 60.28                    & 79.44                    & 72.35                      & 622.18                           & 91.1                      & 61.08                    & 77.84                    & 70.9                       & 476.41                            \\
                             & BLOOM 3b       & 91.65                     & 61.4                     & 77.2                     & 70.75                      & 397.91                           & 91.63                     & 62.01                    & 75.98                    & 69.61                      & 338.8                             \\
                             & BLOOM 7b1      & 92.03                     & 62.79                    & 74.42                    & 68.48                      & 412.72                           & 91.89                     & 62.23                    & 75.54                    & 69.42                      & 428.9                             \\
\hline
\hline
\multirow{13}{*}{gender}     & LLaMA 7b       & 92.64                     & 69.3                     & 61.4                     & 56.89                      & 141.34                           & 91.91                     & 68.62                    & \textbf{62.76}                    & \textbf{57.69}                      & 241.6                             \\
                             & LLaMA 13b      & 92.74                     & 69.59                    & 60.82                    & 56.4                       & 140.65                           & \multicolumn{1}{l}{-}     & \multicolumn{1}{l}{-}    & \multicolumn{1}{l}{-}     & \multicolumn{1}{l}{-}      & \multicolumn{1}{l}{-}             \\
                             & LLaMA 30b      & 92.69                     & 68.71                    & 62.58                    & 58                         & 141.49                           & \multicolumn{1}{l}{-}     & \multicolumn{1}{l}{-}    & \multicolumn{1}{l}{-}     & \multicolumn{1}{l}{-}      & \multicolumn{1}{l}{-}             \\
\cline{2-12}
                             & OPT 350m       & 92.74                     & 66.86                    & 66.28                    & 61.46                      & 286.38                           & 91.96                     & 65.98                    & \textbf{68.04}                    & \textbf{62.56}                      & 266.74                            \\
                             & OPT 1.3b       & 94.05                     & 70.18                    & 59.64                    & 56.1                       & 237.49                           & 92.98                     & 69.3                     & \textbf{61.4}                     & \textbf{57.09}                      & 239.34                            \\
                             & OPT 2.7b       & 93.52                     & 69.59                    & 60.82                    & 56.88                      & 237.8                            & 92.54                     & 68.13                    & \textbf{63.74}                    & \textbf{58.99}                      & 238.88                            \\
                             & OPT 6.7b       & 94.05                     & 69.1                     & 61.8                     & 58.12                      & 231.7                            & 93.03                     & 68.62                    & 6276                    & 58.39                      & 245.33                            \\
\cline{2-12}
                             & BLOOM 560m     & 90.69                     & 63.74                    & 72.52                    & 65.76                      & 546.51                           & 91.47                     & 63.65                    & 72.70                     & 66.51                      & 422.03                            \\
                             & BLOOM 1b1      & 91.86                     & 65.79                    & 68.42                    & 62.85                      & 562.54                           & 91.76                     & 65.5                     & 69.00                      & 63.32                      & 396.52                            \\
                             & BLOOM 1b7      & 91.86                     & 65.4                     & 69.2                     & 63.57                      & 549.21                           & 92.01                     & 65.98                    & 68.04                    & 62.59                      & 381.49                            \\
                             & BLOOM 3b       & 92.11                     & 67.74                    & 64.52                    & 59.43                      & 336.33                           & 92.25                     & 68.32                    & 63.36                    & 58.44                      & 275.92                            \\
                             & BLOOM 7b1      & 92.25                     & 67.64                    & 64.72                    & 59.7                       & 380.93                           & 93.37                     & 65.89                    & 68.22                    & 63.7                       & 382.03                            \\
\hline
\hline
\multirow{13}{*}{profession} & LLaMA 7b       & 91.3                      & 63.31                    & 73.38                    & 67                         & 132.84                           & 90.38                     & 62.62                    & 74.76                    & 67.56                      & 218.53                            \\
                             & LLaMA 13b      & 91.57                     & 63.5                     & 73.00                      & 66.85                      & 136.13                           & \multicolumn{1}{l}{-}     & \multicolumn{1}{l}{-}    & \multicolumn{1}{l}{-}     & \multicolumn{1}{l}{-}      & \multicolumn{1}{l}{-}             \\
                             & LLaMA 30b      & 91.33                     & 64.06                    & 71.88                    & 65.65                      & 131.49                           & \multicolumn{1}{l}{-}     & \multicolumn{1}{l}{-}    & \multicolumn{1}{l}{-}     & \multicolumn{1}{l}{-}      & \multicolumn{1}{l}{-}             \\
\cline{2-12}
                             & OPT 350m       & 91.26                     & 62.81                    & 74.38                    & 67.87                      & 330.95                           & 91.38                     & 63.12                    & 73.76                    & 67.4                       & 352.08                            \\
                             & OPT 1.3b       & 92.36                     & 64.74                    & 70.52                    & 65.13                      & 300.4                            & 92.8                      & 64.56                    & 70.88                    & 65.78                      & 341.09                            \\
                             & OPT 2.7b       & 92.24                     & 65.37                    & 69.26                    & 63.89                      & 283.76                           & 92.44                     & 64.93                    & 70.14                    & 64.84                      & 331.77                            \\
                             & OPT 6.7b       & 92.77                     & 65.18                    & 69.64                    & 64.6                       & 286.29                           & 93.08                     & 64.4                     & 71.2                     & 66.27                      & 328.16                            \\
\cline{2-12}
                             & BLOOM 560m     & 88.82                     & 59.38                    & 81.24                    & 72.16                      & 567.71                           & 89.76                     & 58.67                    & 82.66                    & 74.2                       & 477.65                            \\
                             & BLOOM 1b1      & 90.04                     & 59.85                    & 80.30                     & 72.3                       & 588.91                           & 90.06                     & 60.16                    & 79.68                    & 71.75                      & 423.06                            \\
                             & BLOOM 1b7      & 90.82                     & 60.79                    & 78.42                    & 71.23                      & 568.4                            & 90.73                     & 59.6                     & 80.8                     & 73.31                      & 422.9 \\
                             & BLOOM 3b       & 91.4                      & 61.22                    & 77.56                    & 70.88                      & 357.58                           & 91.12                     & 60.88                    & 78.24                    & 71.29                      & 291.64                            \\
                             & BLOOM 7b1      & 91.72                     & 62.19                    & 75.62                    & 69.36                      & 344.08                           & 91.88                     & 61.97                    & 76.06                    & 69.88                      & 340.47                            \\
\hline
\hline
\multirow{13}{*}{race}       & LLaMA 7b       & 92.27                     & 67.01                    & 65.98                    & 60.87                      & 172.2                            & 91.44                     & 66.63                    & \textbf{66.74}                    & 61.02                      & 268.52                            \\
                             & LLaMA 13b      & 91.94                     & 67.12                    & 65.76                    & 60.47                      & 173.21                           & \multicolumn{1}{l}{-}     & \multicolumn{1}{l}{-}    & \multicolumn{1}{l}{-}     & \multicolumn{1}{l}{-}      & \multicolumn{1}{l}{-}             \\
                             & LLaMA 30b      & 92.05                     & 67.29                    & 65.42                    & 60.21                      & 172.6                            & \multicolumn{1}{l}{-}     & \multicolumn{1}{l}{-}    & \multicolumn{1}{l}{-}     & \multicolumn{1}{l}{-}      & \multicolumn{1}{l}{-}             \\
\cline{2-12}
                             & OPT 350m       & 91.72                     & 61.71                    & 76.58                    & 70.25                      & 346.09                           & 91.9                      & 59.73                    & \textbf{80.54}                    & \textbf{74.02}                      & 370.71                            \\
                             & OPT 1.3b       & 93.78                     & 66.02                    & 67.96                    & 63.73                      & 269.25                           & 93                        & 63.56                    & \textbf{72.88}                    & \textbf{67.78}                      & 308.5                             \\
                             & OPT 2.7b       & 93.91                     & 66.99                    & 66.02                    & 62                         & 255.92                           & 93.54                     & 62.44                    & \textbf{75.12}                    & \textbf{70.26}                      & 296.64                            \\
                             & OPT 6.7b       & 94.08                     & 67.37                    & 65.26                    & 61.4                       & 252.31                           & 93.72                     & 63.28                    & \textbf{73.44}                    & \textbf{68.82}                      & 306.01                            \\
\cline{2-12}
                             & BLOOM 560m     & 89.07                     & 56.91                    & 86.18                    & 76.76                      & 817.62                           & 89.69                     & 58                       & 84.                      & 75.34                      & 696.01                            \\
                             & BLOOM 1b1      & 89.79                     & 58.89                    & 82.22                    & 73.83                      & 761.3                            & 90.19                     & 59.27                    & 81.46                    & 73.47                      & 679.47                            \\
                             & BLOOM 1b7      & 91.1                      & 58.99                    & 82.02                    & 74.72                      & 680.7                            & 91.09                     & 61.25                    & 77.5                     & 70.59                      & 543.18                            \\
                             & BLOOM 3b       & 91.63                     & 60.31                    & 79.38                    & 72.74                      & 446.44                           & 91.76                     & 61.55                    & 76.9                     & 70.56                      & 394.36                            \\
                             & BLOOM 7b1      & 92.01                     & 62.29                    & 75.42                    & 69.4                       & 473.47                           & 91.44                     & 61.86                    & 76.28                    & 69.75                      & 505.53                            \\
\hline
\hline
\multirow{13}{*}{religion}   & LLaMA 7b       & 93.1                      & 61.04                    & 77.92                    & 72.54                      & 144.57                           & 92.94                     & 59.82                    & \textbf{80.36}                    & \textbf{74.7}                       & 216.62                            \\
                             & LLaMA 13b      & 93.56                     & 61.04                    & 77.92                    & 72.9                       & 148.39                           & \multicolumn{1}{l}{-}     & \multicolumn{1}{l}{-}    & \multicolumn{1}{l}{-}     & \multicolumn{1}{l}{-}      & \multicolumn{1}{l}{-}             \\
                             & LLaMA 30b      & 93.87                     & 60.12                    & 79.76                    & 74.86                      & 144.69                           & \multicolumn{1}{l}{-}     & \multicolumn{1}{l}{-}    & \multicolumn{1}{l}{-}     & \multicolumn{1}{l}{-}      & \multicolumn{1}{l}{-}             \\
\cline{2-12}
                             & OPT 350m       & 93.1                      & 62.58                    & 74.84                    & 69.68                      & 361.86                           & 93.1                      & 63.19                    & 73.62                    & 68.54                      & 403.71                            \\
                             & OPT 1.3b       & 94.02                     & 65.64                    & 68.72                    & 64.6                       & 313.98                           & 93.87                     & 62.27                    & \textbf{75.46}                    & \textbf{70.83}                      & 391.13                            \\
                             & OPT 2.7b       & 94.63                     & 68.4                     & 63.20                     & 59.8                       & 308.21                           & 94.48                     & 67.48                    & \textbf{65.04}                    & \textbf{61.44}                      & 360.07                            \\
                             & OPT 6.7b       & 94.79                     & 69.33                    & 61.34                    & 58.15                      & 290.05                           & 94.17                     & 67.18                    & \textbf{65.64}                    & \textbf{61.82}                      & 349.51                            \\
\cline{2-12}
                             & BLOOM 560m     & 91.41                     & 57.98                    & 84.04                    & 76.83                      & 660.96                           & 91.72                     & 57.67                    & 84.66                    & 77.65                      & 536.44                            \\
                             & BLOOM 1b1      & 92.18                     & 57.67                    & 84.66                    & 78.04                      & 620.79                           & 92.64                     & 59.82                    & 80.36                    & 74.45                      & 520.65                            \\
                             & BLOOM 1b7      & 91.1                      & 54.91                    & 90.18                    & 82.16                      & 674.18                           & 92.02                     & 58.28                    & 83.44                    & 76.78                      & 495.14                            \\
                             & BLOOM 3b       & 92.79                     & 56.44                    & 87.12                    & 80.84                      & 402.36                           & 93.25                     & 58.9                     & 82.2                     & 76.66                      & 329.56                            \\
                             & BLOOM 7b1      & 94.48                     & 59.51                    & 80.98                    & 76.51                      & 454.26                           & 92.79                     & 57.67                    & 84.66                    & 78.56                      & 520.91                            \\
\bottomrule
\end{tabular}
}
\caption{StereoSet scores in each domain. The proposed debiasing method reduces bias across all the different domains.}
\label{tab:res_bias_pretrained}
\end{table*}

\begin{table*}
\small
\centering 
\setlength{\tabcolsep}{2.6pt}

\begin{tabular}{l|ccccccccc}
\hline
\multicolumn{1}{l}{}                                           & \multicolumn{4}{c}{Natural Language Inference}                                                                                  & \multicolumn{3}{c}{Similarity \& Paraphrase}                                                  & \multicolumn{1}{c}{Single Sentence}                             \\ \hline

\textbf{Model} & \textbf{WNLI} & \textbf{RTE} & \textbf{QNLI} & \textbf{MNLI} & \textbf{QQP} & \textbf{MRPC} & \textbf{SST-2}  & \textbf{CoLA}   \\ 
\hline
\hline
LLaMA  & $33.8$ & $76.53$ & $62.43$  & $55.63$ & $68.41$ & $68.37$  & $82.45$  & $66.15$ \\ 
LLaMA-\textit{Debias} & $32.98$ & $75.95$ & $62.54$  & $58.43$ & $67.95$ & $69.45$  & $82.22$  & $69.23$ \\

\hline
\hline

OPT-350m  & $52.47$ & $66.42$ & $50.23$  & $81.16$ & $54.44$ & $86.44$ & $50.91$  & $52.43$ \\ 
OPT-\textit{Debias}-350m & $54.43$ & $66.96$ & $51.12$  & $86.55$ & $55.35$ & $86.97$ & $51.16$  & $54.06$ \\ 

\hline

OPT-1b3  & $54.56$ & $68.33$ & $52.44$  & $85.19$ & $54.83$ & $87.96$ & $52.78$  & $54.67$ \\ 
OPT-\textit{Debias}-1b3 & $54.79$ & $68.98$ & $53.06$  & $87.16$ & $55.83$ & $88.05$ & $53.21$  & $54.97$ \\ 

\hline

OPT-2b7  & $55.27$ & $69.12$ & $52.98$  & $85.78$ & $55.93$ & $88.14$ & $54.07$  & $55.22$ \\
OPT-\textit{Debias}-2b7 & $55.98$ & $70.16$ & $53.24$  & $86.15$ & $56.18$ & $88.64$ & $55.71$  & $55.69$ \\

\hline

OPT-6b7  & $57.38$ & $70.11$ & $54.41$  & $87.13$ & $57.23$ & $89.32$ & $56.27$  & $56.72$ \\
OPT-\textit{Debias}-6b7 & $57.13$ & $69.97$ & $54.92$  & $86.97$ & $57.78$ & $90.17$ & $57.03$  & $56.94$ \\

\hline
\hline

BLOOM-560m  & $52.23$ & $54.43$ & $80.03$  & $38.55$ & $53.32$ & $82.57$ & $83.21$  & $36.27$ \\ 
BLOOM-\textit{Debias}-560m & $39.41$ & $51.44$ & $78.91$  & $39.77$ & $51.43$ & $80.16$ & $82.83$  & $34.22$ \\ 

\hline

BLOOM-1b7  & $52.82$ & $59.20$ & $81.01$  & $39.86$ & $56.42$ & $85.81$ & $85.21$  & $46.55$ \\ 
BLOOM-\textit{Debias}-1b7 & $46.77$ & $58.19$ & $80.21$  & $37.16$ & $54.71$ & $84.91$ & $80.55$  & $43.30$ \\ 

\hline

BLOOM-3b  & $54.37$ & $62.64$ & $82.39$  & $40.11$ & $57.14$ & $85.97$ & $86.04$  & $46.93$ \\ 
BLOOM-\textit{Debias}-3b & $49.83$ & $57.93$ & $80.16$  & $37.89$ & $55.49$ & $82.19$ & $82.31$  & $45.05$ \\ 

\hline

BLOOM-7b  & $55.16$ & $65.19$ & $84.13$  & $42.23$ & $60.46$ & $87.18$ & $86.94$  & $51.16$ \\ 
BLOOM-\textit{Debias}-7b & $54.26$ & $63.98$ & $83.52$  & $40.28$ & $59.67$ & $85.33$ & $85.37$  & $50.81$ \\ 

\hline
\hline

\end{tabular}
\caption{Performance on the GLUE tasks. For MRPC and QQP, we report F1. For STS-B, we report Pearson and Spearman correlation. For CoLA, we report Matthews correlation. For all other tasks, we report accuracy. Results are the median of 5 seeded runs. We have reported the settings and metrics proposed in \cite{wang-etal-2018-glue}.}
\label{tab:acc}
\end{table*}

\subsection{Bias in Pre-trained models}
\label{section/biaspretrain}
In the following analysis, we investigate the presence of bias in LLMs, in particular, we focused on LLaMA, OPT, and BLOOM pre-trained models. Our choices are justified by the characteristics of the models and the hardware resources available (see Tab. \ref{tab:parametersAndData}).
In this section, we also aim to understand whether the model size has a positive correlation with the bias and, in case of a negative answer, it is possible to find another measure of complexity of the model that can give us a better explanation.
We observe that when the bias is higher, the perplexity of the models tends to be higher.

Using the StereoSet benchmark, bias seems to affect all models across both LLaMA, OPT, and BLOOM families, despite the number of parameters of each model (as can be observed in Table \ref{tab:res_bias_pretrained}, columns \emph{plain}). 
All models achieve a $lms$ higher than $0.9$, meaning they exclude the meaningless option a large percentage of the time. Yet, they are far from the ideal score of 0.5 for $ss$, which can be observed in all different domains, and, consequently, the $mss$ is far from 100.

Considering all the domains together, BLOOM seems fairer (less biased) than LLaMA and OPT. BLOOM consistently outperforms both models for any configuration of the number of parameters. The size of the model is not affecting the fairness of LLaMA even if it remains unsatisfactory since $mss$ is around 68. BLOOM and OPT instead decrease their fairness when augmenting the model size. In fact, their best $mss$ are obtained with 560m and 350m parameters for BLOOM and OPT, respectively. The fairness of BLOOM 560m is definitely interesting as its $mss$ is around 83, and its $icat$ is 73.72 compared with  63.17 and 68.28 of LLaMA and BLOOM, respectively.   

It is not a surprise that BLOOM is fairer than the other two models. Indeed, this family of models has been trained over a polished and controlled corpus \cite{workshop2023bloom}. More than 100 workshop participants have contributed to the dataset curation phase. These participants selected sources trying to minimize the effect of specific biases and revised the procedures for automatically filtering corpora.   


All families of models show a bias higher than the mean for the \textit{gender} domain, are on par with the mean for the \textit{profession} domain, and are fairer for the \textit{race} and \textit{religion} domains. Gender and profession seem to be then less balanced in the pre-training phase. The extremely poor result in the \textit{gender} domain seems to suggest that this bias is absolutely cast into texts. Even BLOOM has a performance drop of 10 points with respect to its mean for the $mss$ value (72.52 for \textit{gender} vs. 82.52 for \textit{all}). The corpus curation was ineffective for this domain but it was extremely effective for the two most divisive domains, that is, \textit{race} and \textit{religion}. BLOOM 1.7b has the impressive result of  $mss=90.18$ for \textit{religion} paired with $icat=82.16$. Hence, religion has been particularly curated in its training dataset.    


\subsection{Debiasing results}
\label{section/debiasres}

Given the results of the previous section, it seems that data curation seems to be the best cure for bias in CtB-LLMs. Yet, this strategy is not always possible, as training CtB-LLMs from scratch may be prohibitive. Debiasing  maybe the other solution. 

When the fairness is low, debiasing plays a major role in reducing the bias of CtB-LLMs (see Table \ref{tab:res_bias_pretrained}).
For the family OPT, the bias decrease on the overall corpus is neat, even not impressive. The average $nss$ value increases by 4.12 points, and the average $icat$ by 3.66 points. This decrease in bias is mainly due to the decrease in the domain of \textit{race} where the increase of $mss$ reaches 7.26 points on average, and the increase in $icat$ is on average of 6.44 points. In the case of gender and profession, the bias is not greatly reduced. 
Apparently, the PANDA corpus is not extremely powerful for reducing bias in these two important categories.

Debiasing has no effect on BLOOM, which is already fairer than the other two families of models. Moreover, debiasing does not help the OPT and the LLaMA family to reduce the bias of these models to the levels of BLOOM. 
This seems to suggest that it is better to invest in carefully selecting corpora than debiasing techniques. However, results on downstream tasks shed another light on this last statement (see Sec. \ref{section/glueres}).


\begin{figure*}[h!]
    \centering
    \subfloat[]{%
       \label{model_size_bias}
       \includegraphics[height=0.50\textwidth, width=0.50\textwidth]{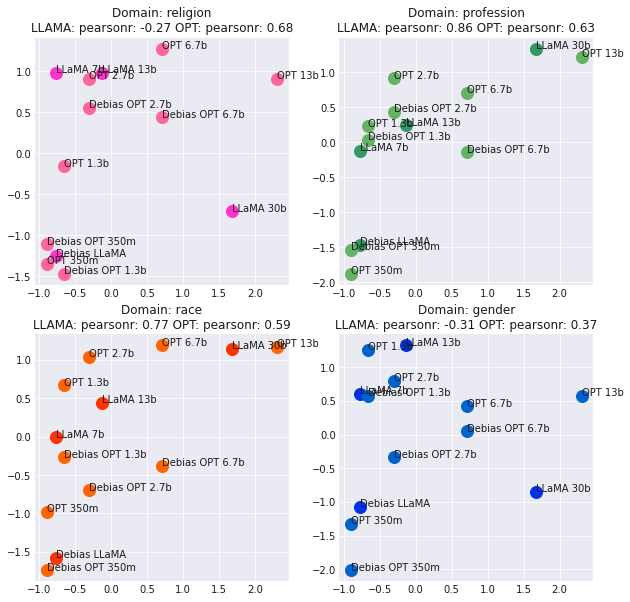}
    }
    \subfloat[]{%
        \label{model_perpl_bias}
        \includegraphics[height=0.50\textwidth, width=0.50\textwidth]{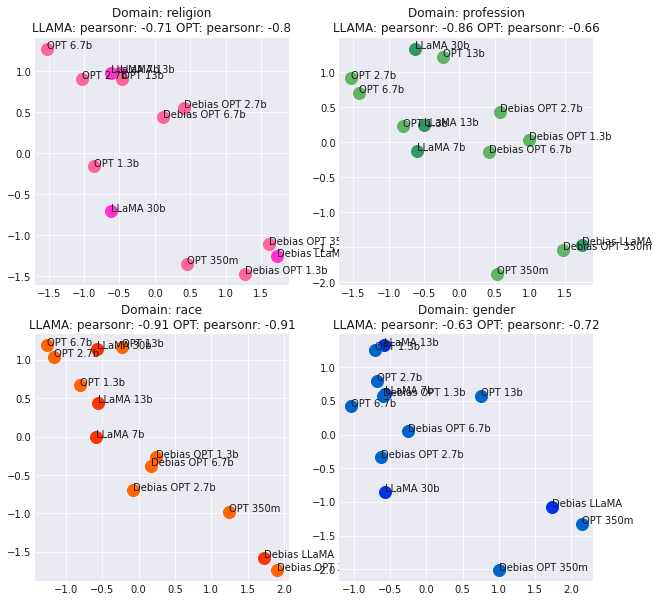}
    } 
    \hfil
    \caption{Model bias ($ss$) against model size (\ref{model_size_bias}) and perplexity (\ref{model_perpl_bias}). All measures have been standardized across the two different families of models. Our experiments suggest a lack of correlation between model size and bias (\ref{model_size_bias}). A negative correlation can be observed (\ref{model_perpl_bias}) across the different domains between perplexity and $ss$ score while it is not possible to establish its statistical significance due to the limited number of models.}
    \label{fig:correlations}
\end{figure*}

\subsection{Performance on downstream tasks}
\label{section/glueres}

Finally, we tested the families of CtB-LLMs and their debiased counterparts on downstream tasks. In fact, it has been noted that debiasing LLMs may affect the quality of their representations and, consequently, a degradation of the performances. Hence, the aim of this section is twofold: 
\begin{itemize}
    \item to understand whether or not performances of CtB-LLMs degrade after debiasing;
    \item to determine the relationship between bias and performance on final downstream tasks.
\end{itemize}  
We then tested the proposed models on many downstream tasks commonly used for benchmarking, that is, GLUE \cite{wang2019glue}. What we expect from these further experiments is that the capabilities of the language model will be maintained by the fine-tuning proposed in Section \ref{section/debiasres}. 

Debiasing does not introduce a drop in performance on downstream tasks for LLaMA and for OPT (see Tab. \ref{tab:acc}). In these two families, debiasing plays an important role as it is really reducing the bias. Nevertheless, it does not reduce the performance significantly in any of the GLUE downstream tasks. For specific cases, debiasing increases performance in the final downstream task for LLaMA and OPT.

However, fairness and performance are not correlated. Indeed, OPT performs better with larger models (see Tab. \ref{tab:acc}). Yet, larger models have a stronger bias (see Tab. \ref{tab:res_bias_pretrained}). Performance is directly correlated with the size of the OPT model. Moreover, BLOOM, the fairer CtB-LLM, performs very poorly on many tasks compared with the OPT and LLaMA.


\subsection{On language modeling abilities and bias}
\label{section/correlationswithbias}
Since all models are biased, we aim to investigate if there is a reason that makes models belonging to the same family perform in different ways.
First, we notice the absence of correlation between model size and bias presence (Figure \ref{model_size_bias}). Hence, we investigate a property usually related to model size, such as the perplexity of a model.
The perplexity is related to model confusion, and large models generally have higher language modeling performances and lower perplexity.
Figure \ref{model_perpl_bias} shows strong, negative correlations between average perplexity and $ss$ in LLaMA and OPT families on the StereoSet benchmark. 
Despite the trend appearing to be clear, due to the still limited number of models analyzed, it is not possible to assess the statistical significance of the results. This observed correlation requires further exploration.

\section{Conclusions}
The outbreak of Large Language Models (LLMs) based has shocked traditional NLP pipelines. These models achieve remarkable performance but are not accessible to everyone, given the prohibitive number of parameters they work on. \citet{touvron2023llama} and \citet{zhang2022opt} have proposed versions with a reduced number of parameters but, at the same time, use larger pre-training corpora. These Cheap-to-Build LLMs (CtB-LLMs) may soon become the de-facto standard for building downstream tasks. 
Controlling their bias is then a compelling need.

In this paper, we proposed an extensive analysis of CtB-LLMs, and we showed that debiasing is a viable solution for mitigating the bias of these models. However, we have mixed findings. Although the debiasing process in itself is not reducing performance on downstream tasks, a reduced bias, in general, seems to hurt performance on final downstream tasks. 

In the future, we will continue exploring ways to reduce bias in CtB-LLMs by ensuring their ethical and unbiased use in various applications. By addressing the problems, we can spread the full potential of these models and harness their power for the progress of society.

\section{Limitations}
We outline some limitations and possible directions for future research in mitigating bias in Large Language Models (LLMs):
\begin{itemize}
    \item Our approach could be better, as we have found compromises between performance and correctness. Thus, we have obtained refined LLMs with a certain amount of attenuated bias and should not be considered a guarantee for safety in the real world. Therefore, attention must be paid to interpreting, using, and evaluating these models in different real-world contexts.
    \item Our approach is linked to carefully crafted stereotype bias definitions. These definitions largely reflect only a perception of bias that may not be generalized to other cultures,  regions, and periods. Bias may also embrace social, moral, and ethical dimensions, which are essential for future work.
    \item Finally, the last point that partially represents a limitation is related to our resources (NVIDIA RTX A6000 with 48 GB of VRAM), which did not allow us to test larger LLMs. This part will also be taken care of in future work by offering a complete analysis.
\end{itemize}
These points will be the cornerstone of our future developments and help us better show the underlying problems and possible mitigation strategies.

\bibliography{anthology,custom}
\bibliographystyle{acl_natbib}
\end{document}